\documentclass[letterpaper]{article}
\usepackage{iccc}

\usepackage{times}
\usepackage{helvet}
\usepackage{courier}
\usepackage{graphicx}
\graphicspath{ {./figures/} }
\usepackage{adjustbox}
\usepackage{url}
\usepackage{natbib}
\usepackage{csquotes}
\usepackage{setspace}
\usepackage{flushend}  % Balanced columns on last page

\usepackage[nolist,nohyperlinks]{acronym}

\newcommand{\point}[1]{\par\noindent \raisebox{.3ex}{$\bullet$}\hskip 0.05in #1}

\title{Automating Generative Deep Learning for Artistic Purposes:\\Challenges and Opportunities}
\date{}

\author{
    Sebastian Berns\,\textsuperscript{1}{\rm ,}
    Terence Broad\,\textsuperscript{2,\,3}{\rm ,}
    Christian Guckelsberger\,\textsuperscript{4,\,5,\,1} \and 
    Simon Colton\,\textsuperscript{1,\,6}\\
    \textsuperscript{1}\,School of Electronic Engineering and Computer Science, Queen Mary University of London, UK\\
    \textsuperscript{2}\,Department of Computing, Goldsmiths, University of London, UK\\
    \textsuperscript{3}\,Creative Computing Institute, University of the Arts London, UK\\
    \textsuperscript{4}\,Department of Computer Science, Aalto University, Espoo, Finland\\
    \textsuperscript{5}\,Finnish Center for Artificial Intelligence\\ 
    \textsuperscript{6}\,SensiLab, Faculty of IT, Monash University, Melbourne, Australia
}

\begin{document} 

\maketitle

\begin{abstract}
We present a framework for automating generative deep learning with a specific focus on artistic applications. 
The framework provides opportunities to hand over creative responsibilities to a generative system as \emph{targets} for automation. For the definition of targets, we adopt core concepts from automated machine learning and an analysis of generative deep learning pipelines, both in standard and artistic settings. 
To motivate the framework, we argue that automation aligns well with the goal of increasing the \emph{creative responsibility} of a generative system, a central theme in computational creativity research. 
% We frame the interaction between the engineer and the generative system as a co-creative process. In this setting, automation becomes the challenge of granting the system more \emph{creative autonomy}. 
% By framing the interaction between the engineer and the generative system as a co-creative process, we define automation as the challenge of granting the system more \emph{creative autonomy}. 
We understand automation as the challenge of granting a generative system more \emph{creative autonomy}, by framing the interaction between the user and the system as a co-creative process. 
% We analyse the relationship between automation and creative autonomy in the context of our framework.
The development of the framework is informed by our analysis of the relationship between automation and creative autonomy.
An illustrative example shows how the framework can give inspiration and guidance in the process of handing over creative responsibility. 
\end{abstract}

\section{Introduction}

The increasing demand in industry and academia for off-the-shelf \ac{ML} methods has generated a high interest in automating the many tasks involved in the development and deployment of \ac{ML} models. Such \ac{AutoML} can make \ac{ML} more widely accessible to non-experts, and decrease the workload in establishing \ac{ML} pipelines, amongst other benefits.

%In the evolution of \ac{DL} as \ac{ML} subdomain, some of the early advances consisted of incorporating previously manual pre-processing steps into an automated optimisation pipeline. While some \ac{ML} algorithms require the extraction and construction of features by hand and image filters have to be selected depending on the task, modern \ac{DL} techniques have no issues with handling raw data, and image filters can easily be learned end-to-end. This brings significant speedups, allowing to scale up the purely computational processes. This has contributed to the consistent successes of the field. As deep learning has become increasingly complex and impactful on society, it has become more pressing and difficult to solve existing challenges, such as the choice of network architecture and the tuning of hyper-parameters.
\ac{AutoML} is a very active area of research. The progress to date has been documented in several surveys \citep[e.g.][]{truong2019towards,tuggener2019automated,chauhan2020automated,he2021automl}. There exist a book \citep{hutter2019automated}, an \ac{AutoML} challenge \citep{automlchallenges} and a dedicated workshop at the International Conference on Machine Learning, currently in its seventh edition. Crucially though, the automation of generative \ac{DL} as \ac{ML} subdomain has received very little attention.

While \ac{AutoML} is concerned with automating solutions for classification and regression, methods in generative \ac{DL} deal with the task of distribution fitting, i.e.~matching a model’s probability distribution to the (unknown) distribution of the data. \Ac{NAS}, an important topic of research in \ac{AutoML}, has been extended to \acp{GAN} \citep{Gong_2019_ICCV,Li_2020_CVPR,Gao_2020_CVPR,pmlr-v119-fu20b}, one prominent type of generative models. Moreover, evolutionary approaches have been applied to optimising the \ac{GAN} training objective \citep{wang2019evolutionary} and other training parameters \citep{costa2019coegan}. Even though certain aspects of the \ac{GAN} training scheme have been automated, we highlight three gaps in existing research: (i) there exists no unified automation framework for generative \ac{DL} more generally; (ii) existing work does not address the use of generative \ac{DL} for artistic purposes; (iii) researchers have not sought to motivate the automation of \ac{DL} systems with the goal to endow artificial systems with creative autonomy.

We propose a framework for the automation of generative deep learning that, on the one hand, adopts core concepts from \ac{AutoML}, and on the other hand, is informed by the theory and practice of \ac{CC} research, the “philosophy, science and engineering of computational systems which, by taking on particular \emph{responsibilities}, exhibit behaviours that unbiased observers would deem to be creative”~\citep[][emphasis added]{colton2012computational}. We can leverage insights from \ac{CC} because automation in generative \ac{DL} aligns with one of the field's central research goals: to endow computational systems with \emph{creative responsibilities}~\citep{colton2009seven}, i.e.~the ability to make specific decisions in a creative process. 
These decisions independently can be understood as \emph{targets} for automation when framing the design of a generative \ac{DL} pipeline as a form of \emph{co-creativity} \citep{kantosalo2014isolation}. 
By virtue of this interpretation, we can inform the automation of generative \ac{DL} more specifically with well-established, generic \ac{CC} strategies to equip computational systems with creative responsibilities. 
Our framework differs from \ac{AutoML} not only in its stronger focus on generative models, but also in the assumed goals of the generative \ac{DL} pipeline. More specifically, we identify targets for automation based on the wide and successful application of generative \ac{DL} in artistic work. 
% for its abundance of examples and. 
In contrast to standard applications, artistic \ac{ML} engineers and users aim to produce artefacts of high cultural value over perfectly generalised reproductions of the training data.
% While the proposed framework is applicable to other instances of creative autonomy, 

Our main contribution is to gather, standardise and highlight opportunities to automate generative \ac{DL} for artistic applications. We identify commonalities of \ac{DL} pipelines in artistic projects and bring them together in a common framework. This provides a starting point for handing over creative responsibilities in a range of applications, not only artistic.
% The presented targets for automation highlight opportunities and challenges.
We concentrate our efforts on generative deep learning, rather than generative \ac{ML} more generally. 
While we assume the majority of applications to be built on \ac{DL} approaches, we do not rule out that other generative \ac{ML} methods might be used within the framework.
Our contribution does not consist of a formal solution to a singular automation problem. In contrast, we aim to provide a big picture view of all automation tasks and their associated opportunities and challenges, to be solved in future work.

To leverage insights from \ac{CC} in the development of our framework, we first clarify the relationship between automating generative \ac{DL} and endowing artificial systems with creative responsibility. %We start by framing the interaction between the engineer and the \ac{DL} system as a co-creative process. 
% In this setting, automation then becomes the challenge of giving more creative autonomy to the \ac{DL} system.
We then outline a standard non-automated pipeline for the development and deployment of generative deep learning models, and show %, with the help of examples from the literature and artistic projects, 
how applications in artistic settings differ from this standard pipeline. Drawing from these two sources, we lay out the automated generative deep learning pipeline, describe several targets for automation therein and suggest ways in which automation could be achieved. 
We continue with an illustrative example to demonstrate how our framework can give inspiration and guidance in the process of gradually handing over creative responsibility to a generative system. 
% A demonstrative example, which shows the effects that automating individual targets can have, is left for future work.
We analyse the relationship between automation and creative autonomy in the context of our framework.
We conclude the paper by discussing the limitations of our framework and suggest directions for future work.

% We aim to increase bridging between deep learning and \ac{CC} research \citep{berns2020bridging}. In particular, it is second nature for \ac{CC} researchers to consider handing over creative responsibilities to software. However, this may involve the identification and breaking of some assumptions around human control for deep learning researchers, and we hope the framework described here provides an encouraging and systematic approach for doing this.

\section{Automated, Artistic Deep Learning \\as Co-Creation}

We believe that the development of a framework for automated generative \ac{DL} can benefit from the insights gathered over more than two decades of \ac{CC} research, because the automation of targets in generative \ac{DL} can be considered a specific instance of the grand \ac{CC} goal to give computational systems responsibility over decisions in a creative process.% \citep{colton2009seven}.

With each creative responsibility that is handed over to the system, i.e.~with each target that is being automated, we increase the computational system’s \emph{creative autonomy} \citep{jennings2010developing,mccormack2019autonomy, Guckelsberger2017a}, i.e.~its capacity to operate independently of a human instructor, allowing for it to be ultimately considered a creator in its own right~\citep{colton2008creativity}. Crucially though, the users of automated generative \ac{DL} typically want to retain some control over the automation and its outcome. In developing our framework, we must thus decide which responsibilities should be retained in order to sustain certain modes of interaction between the artistic users and the generative \ac{DL} system. 

To this end, it is useful to frame this interaction in the process of automation as a \emph{co-creative} act. We adopt Kantosalo et al.'s \citeyearpar{kantosalo2014isolation} working definition of \emph{human-computer co-creativity} as \enquote{collaborative creativity where both the human and the computer take creative responsibility for the generation of a creative artefact}. To qualify as a collaborative activity, both human and system must achieve \emph{shared goals} (Kantosalo et al., \citeyear{kantosalo2014isolation}, drawing on \citeauthor{terveen1995overview}, \citeyear{terveen1995overview}). 

Different automation strategies can enable two coarse forms of interaction. 
First, the user and system could engage in \emph{task-divided co-creativity}, in which \enquote{co-creative partners take specific roles within the co-creative process, producing new concepts satisfying the requirements of one party} \citep{kantosalo2016modes}. 
Second, they could engage in \emph{alternating co-creativity}, where both partners \enquote{take turns in creating a new concept satisfying the requirements of both parties} \citep{kantosalo2016modes}.

Alternating co-creativity requires the computational system to not only exhibit creative responsibility for either the \emph{generation} or \emph{evaluation} of artefacts, but for both. Crucially, even a non-automated generative \ac{DL} system can be considered creative in a minimal sense, in that it (despite the name) not only \enquote{merely \emph{generates}} \citep{ventura2016mere} new samples or artefacts, but also \emph{evaluates} their proximity to the training set via its loss function. This is accomplished either explicitly, through likelihood estimation, or implicitly, with the help of a critic in an adversarial setting. The system thus produces artefacts that are \emph{novel} and \emph{valuable}, realising both requirements of the two-component standard definition of creativity \citep{runco2012standard}. We write \enquote{creative in a minimal sense}, because the novelty of artefacts will decline, while their value increases, the better the system approximates the (unknown) distribution from which the training data was drawn. 
% This could be counteracted by architectures or objective functions that also optimise for novelty, e.g.~creative adversarial networks~\citep{elgammal2017can}.

The definition of the training set and loss function by the user satisfies that both partners interact towards shared goals. Through different ways to automate the \ac{ML} pipeline, we can free the human partner from certain manual work, while retaining specific creative responsibilities. 
% We later distinguish possible automation variants by mapping them to the modes of \emph{task-divided} and \emph{alternating co-creativity}.

We believe that providing the computational system with creative responsibility in the form of automating certain targets does not constrain, but rather expands the shared creative process. The \emph{person} or \emph{producer} has, due to their personality and cognitive characteristics, a strong impact on the creative \emph{process}, \emph{product}, and the creative environment, i.e.~the \emph{press}~\citep{rhodes1961analysis, jordanous2016four}. However, human creativity is also limited, e.g.~due to our bounded rationality~\citep{simon1990bounded}. A computational system can complement human shortcomings, e.g.~via its higher information processing or memory capacity, enabling creativity on larger search spaces~\citep{boden2004creative, wiggins2006searching}.

%The arguments for Creativity autonomy are insightful but by providing AutoML are we unlocking or constraining the Creative process? Fewer variables, less freedom for the system and the tendency of co-creation systems tend to be taylorMade solutions that unlock the creative process of the user and or machine. The proposal tends to lock it down to data and some parameters to control training and evaluation. It would be interesting to reflect on these aspects in this article.
%Potentially argue with 4 P: person puts constraints on process, too (e.g. bounded rationality, time/resource constraints).

% \notes{Sebastian: potentially highlight additional means to include generation and evaluation in the system as part of automation later. Not strictly necessary, but potentially of interest}. 

\section{A Standard Generative Pipeline \\and Artistic Deviations}

We outline the various steps in the process of building and deploying a generative \ac{DL} model for standard non-automated usage and contrast it with the particular differences that arise when using a model in different artistic contexts. 
% For example, it is much more common in a creative context to make iterative modifications to the training data set, rather than the model architecture and training parameters.
Additionally, we provide a brief overview post-training modifications that aim for active divergence \citep{berns2020bridging}, allowing to manipulate a model into producing artefacts that do not exactly resemble the training data. A more detailed survey of such techniques can be found in \citet{broad2021active}.
Our goal is to highlight the many choices that have to be taken in the construction of a generative \ac{DL} pipeline and identify those tasks which pose an opportunity for automation in our framework.

\subsection{Data Acquisition}

The first step towards developing generative models is data acquisition. We distinguish two cases: (i) using pre-existing data sets and (ii) creating new ones. It should be noted, that generative \ac{ML} is also applied in privacy sensitive areas such as medicine, and in the augmentation of small data sets, as it can produce synthetic data to replace an entire data set or supplement it with additional samples. The augmentation by way of a generative model can be necessary whenever a data set is too small to train another model (e.g.~a classifier) with a high number of parameters (i.e.~weights and biases in a neural network). However, when the generative model itself requires a large amount of training data, other pre-training data augmentation steps through graphic manipulations can help to do so effectively \citep{karras2020training}. 

\subsubsection{Using Existing Data Sets}

In a research setting, it is most common to use standard benchmark data sets or subsets thereof, for training and evaluating generative models. It is generally best practice in machine learning to split the data into training, test and validation subsets. However, generative models are sometimes trained on the entire data set and alternative methods of evaluation are used.

\subsubsection{Creating a New Data Set}

When creating a data set from scratch, the goal is normally to fully represent the subject or category that is being modelled. Therefore, as much data as possible will be collected to maximise variation in the data set and to represent all modes as evenly as possible, i.e.~the variety of artefacts that are statistically significantly different from one another. 
Creating varied, high-quality data sets with the large amounts of data required for training generative models can be very labour intensive and usually the purview of a select few academic and industry laboratories. This is often performed in a distributed fashion, where many workers are involved in collecting, evaluating and labelling data samples. 

% \notes{Artistic deviation}
In contrast to data sets created for industrial and research applications, data sets for artistic purposes are often composed with very different goals. It may not be important to accurately and fully represent a subject matter or domain, as long as the end goal produces interesting results. Data sets are often much smaller, and the considerations for the desired aesthetic characteristics in the end results are much more important in deciding which examples should and which should not be included in the data set. A lot of effort will go into sourcing material and the resulting data sets are much more likely to be reflect an artists individual style and (visual) language. In some cases, the entire data set will come from an artist’s personal archive \citep{ridler2017repeating}.

\subsection{Training}

The objective of training a generative model is to learn a mapping function from an easily controllable and well understood distribution, e.g.~a standard Gaussian, to a distribution of much higher complexity and dimensionality, e.g.~that of natural colour images. 
% Elaborate on different architecture types (explicit, implicit) and training schemes
There are a number of different training schemes, which apply to different architectures. They are commonly categorised by their formulation of the training objective. Methods maximise the likelihood of the data either (i) explicitly, such as auto-regressive and flow-based models, (ii) approximately, e.g.~\acp{VAE}, or (iii) implicitly (\acp{GAN}). When using a method that explicitly models the data, training will be performed until a desired likelihood score is reached. With \ac{VAE}s, the goal of training is to maximise the log-likelihood of the data set. In the adversarial setup, the decision when to stop training is less clear. Training is often run for a pre-specified period and the results are evaluated qualitatively.
A fully trained model ideally represents the entire training data distribution, and can be sampled randomly to produce good results. Another desirable quality is that interpolation between two input vectors is matched in the outputs.

% \notes{Artistic deviation}
Generalisation is a goal of almost all \ac{ML} systems and applications. A model should be able to generalise to unseen data, while not underfitting or overfitting the training data. In an artistic setting, however, this is often less important, and if it produces interesting results, artists may often embrace the aesthetic qualities of an underfit \citep{shane2018machine} or overfit model \citep{broad2017autoencoding}.

\subsection{Evaluation}

The general performance of a model is measured in terms of the distance of the learned distribution to the target distribution. A model further ideally covers all modes in the input data set.
For generative methods that explicitly model a probability distribution over the data, the (log) likelihood can be measured and evaluated directly. Implicit methods, such as \acp{GAN}, have to be assessed with other metrics such as the Inception Score \citep{salimans2016improved} and the \ac{FID} \citep{heusel2017gans}. As these metrics are only a simplified standard for evaluation and have some shortcomings, additional qualitative checks might be needed to ensure fidelity of the output. 

% In the standard setting the quality of an output sample is usually defined as, whereas diversity refers to the breadth of represented modes. From an artistic point of view, however, the understanding of quality can differ significantly from this standard definition and 

While in some artistic settings good quantitative performance might matter, it can be ignored entirely in others, and a qualitative assessment of the output is usually much more important. 
% In addition to a quantitative assessment of a model, artistic evaluation of a model is primarily done qualitatively, rather than quantitatively, and the desired outcomes from the qualitative assessment may be very different from that of researchers. 
Quality, diversity and accuracy may not be the only considerations (and may even be actively avoided), whereas novelty, interesting mis-representations of the data and other aesthetic qualities may be desired. 
Due to the variety of qualities that an artist might look for in a model’s output, there is no unique or widely used standard metric for evaluation. This is rooted in the highly individualistic nature of artistic work and linked to the additional strategies for iterative improvements and curation of the output which we discuss in the following subsections.
% Unknown unknowns, looking for that which we did not know we were looking for

\subsection{Iterative Improvements of Outputs}

Here we look at the diverging strategies for the gradual improvement of a system’s output in a research and development versus an artistic setting. 
% From the perspective of the interaction between human artist and generative \ac{DL} system, as we discussed above, this can be considered an alternating co-creative process.

\subsubsection{Iterating on the Model}

In the research and development of generative models, the data set often remains fixed, while various aspects of the network architecture and training regime will be altered. For instance, various optimisation hyper-parameters will be evaluated, such as: learning rate, momentum or batch size; or network configurations: number of layers, type of activation functions, etc. Different training regimes may also be experimented with, such as: optimisation algorithms, loss functions, and methods for regularisation and sampling. 

\subsubsection{Iterating on the Data Set}

In artistic contexts, it is much more common to iterate on the data set and keep other parameters fixed, before possibly making iterative improvements to the network and model parameters. Data that appears to be producing unwanted results, or skewing the model in certain directions may be removed. Revisiting the composition of samples (such as cropping), and the removal and addition of samples in order to refine the data set may be undertaken \citep{schultz2020datasets}.

\subsection{Deployment}

Generative models are used differently in standard and artistic settings in accordance with their respective goal. We here differentiate between standard sampling and output curation.

\subsubsection{Standard Sampling}

Generative models are trained with the goal that they can be sampled randomly and every generated output will be of value and high typicality \citep{ritchie2007empirical}. Therefore, in most standard applications models are simply sampled randomly with no additional filtering taking place. When filtering is performed, it is often done with the goal of quality evaluation, such as using the discriminator for evaluation quality \citep{azadi2019discriminator}, or using the \ac{CLIP} model \citep{radford2021learning}, as was the case in evaluating and ranking the generated outputs of the discrete \ac{VAE} model in the DALL-E image generation project \citep{ramesh2021zero}.

\subsubsection{Output Curation}

Rather than sampling randomly from a model, artists will often spend a lot of time curating a model’s output. The goal of building a model in an artistic setting is not necessarily to generate only samples of high value, but to produce some interesting or novel results, which can then be hand-selected. This can be through filtering samples or searching and exploring the latent space. In some cases, such as combining language-image models with latent space search for text-to-image generation, e.g.~\citet{murdock2021big}, much effort goes into prompt engineering to find a specific latent vector that produces interesting results.

\subsection{Post-training Modifications}

Having looked previously at the curation of a model’s output in an artistic setting, i.e.~the act of identifying the few artefacts of interest in a large set of output samples, we now turn to active divergence techniques \citep{berns2020bridging} which aim at consistently producing results that diverge from the training data.
These strategies, specifically developed in creative contexts for the purpose of art production, include hacks, tricks and modifications to the model parameters, as well as the daisy-chaining of several models.

One approach is to find a set of parameters where the generated artefacts blend characteristics of multiple data sets. 
For this, a pre-trained model can be fine-tuned on a second data set, different from the original data. As soon as the results present an optimal blend between the two data domains, the fine-tuning can be stopped. This mixture of data sets can also be achieved by blending the weights of two models. Either interpolating on the weight parameters of the two models, or swapping layers between models, so that the new model contains higher level characteristics of one model, and lower level characteristics of another. 
Another method consists in chaining multiple models together. This allows artists to explore and combine characteristics of different data sets. Unconditional generative models will often be chained together with domain-translation models, e.g.~CycleGAN \citep{zhu2017unpaired} for sketch-to-image translation, or style transfer algorithms \citep{gatys2016image}. The aim of such pipelines is to produce artefacts that reflect the complex combination of characteristics from many data sets. 

Other approaches make modifications to the model in order to have artefacts completely diverge from any training data. An existing pre-trained model can be fine-tuned using a loss function that maximises the likelihood over the training data \citep{broad2020amplifying}. %, an approach referred to as \textit{loss hacking}. 
Other techniques intelligently combine learned features across various models \citep{guzdial2018combinets}, or rewrite the weights of the model \citep{bau2020rewriting}, re-configuring them to represent novel data categories or semantic relationships. In contrast, \emph{network bending} does not require any changes to the weights of the model \citep{broad2021network}. An analysis of the model is performed to determine which features are responsible for generating different semantic properties in the generated output. Deterministically controlled filters are then inserted as new layers into a model and applied to the activation maps of features.

\vskip 0.9cm

\section{An Automation Framework}

\begin{figure*}[!htb]
    \centering
    \includegraphics[width=0.72\linewidth]{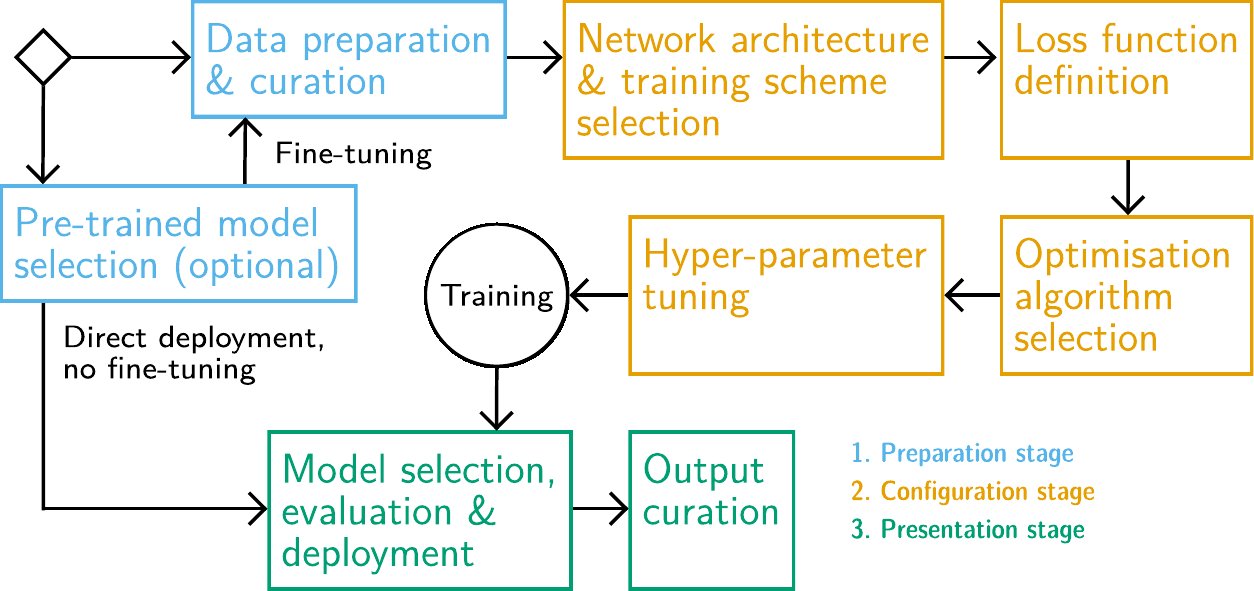}
    \caption{Automated generative deep learning framework. Three stages: preparation, configuration and presentation. Targets for automation are illustrated as individual steps and are further discussed in the section \emph{Targets for Automation} below.}
    \label{fig:pipeline}
\end{figure*}

We build our framework drawing on the standard generative \ac{DL} pipeline and its artistic deviations, as previously described. 
% The important difference between the pipeline for a standard generative model and its use in a creative setting is the overall objective of the project. While the former focuses on the model and its performance in terms of fitting the data distribution as closely as possible, a good model in the context of the latter is one that can reliably produce output of high cultural value. We therefore do not simply extend core concepts of {AutoML}, but join them with practices observed in the wild and re-frame them within the paradigms of \ac{CC}.
We first discuss automation as a search problem and the various techniques it can be approached. We then go on to list the targets for automation in a generative deep learning pipeline for artistic purposes. Some decisions have implications for other targets further down the line, e.g.~the number and type of hyper-parameters depend in part on the kind of network architecture and optimisation algorithm. While the process is presented as a sequence of consecutive steps from input to output, it should be understood that all steps are optional and flexibility is required. Improving a system’s output works best as an iterative loop in which we might go back and adjust or intervene at any given prior step.

% \subsection{Terminology}

We define the terminology of our framework as follows. With \textit{automation}, we refer to the act of addressing with computational means those decisions in a generative deep learning pipeline that normally would be taken by a person. 
A \textit{target} is defined as one such decision which provides an opportunity for automated instead of manual tuning.
% We further refer to \textit{meta-parameters} as the set of values of all targets, i.e.~the specific configuration of a pipeline. This stands in contrast to the model parameters and the hyper-parameters, which define the configuration of a target, e.g.~the optimisation algorithm.
% We define the optimal system as a generative deep learning pipeline which produces as output an artefact of high cultural value.

\subsection{Automation as a Search Problem}

% For illustrative purposes we here refer to targets as \textit{meta-parameters}, i.e.~the specific configuration of a pipeline. This stands in contrast to the model parameters and the hyper-parameters, which define the configuration of a target, e.g.~the optimisation algorithm.
A generative pipeline is automated by assigning responsibilities over individual targets to either the user or the system. While those retained by a person will have to be tuned manually, all other targets require the system to determine a configuration independently. This problem is analogous to the search problem over hyper-parameters in \ac{AutoML}. The possible values of each automated target effectively construct a search space over possible system configurations. The number of total permutations, and the resulting search space, can grow rapidly with every independent target added.
 
Limiting continuous parameter values to a reduced range or a set of discrete values, as per grid search for machine learning hyper-parameters, can help make the problem more feasible. The formulation as a search problem is the standard way to tackle automation in {AutoML}. However, extensive search over meta-parameters can be computationally expensive, time-consuming, cause high energy consumption and consequently have a considerable environmental impact.

The extensive work on search problems provides numerous approaches to constrain this search. Strategies range from complete, to informed, to random methods. While exhaustive search can yield an optimal solution, it can be impractical and often infeasible for large search spaces. 
Random sampling, on the other extreme, can be a surprisingly effective strategy at low cost and with potentially surprising results. 
While \citet{jennings2010developing} requires a system to meet the \emph{non-randomness} criterion in order to be considered creatively autonomous, this definition does not rule out all uses of randomness and allows for testing random perturbations to a system’s standards.
AI-based search methods can benefit from meaningful heuristics and leverage both exploration and exploitation (e.g.~evolutionary search). Gradient-based methods have seen a lot of progress in recent years. Other approaches include rule-based selection and expert systems, with drawbacks including that they require manual construction and expert knowledge. 

Finally, machine learning itself can be used to choose values through a pre-trained model.
Indeed, practitioners in generative deep learning tend to go directly to automation via deep learning. In particular, recent advances in contrastive language-image pre-training \citep{radford2021learning} allow for computing similarities between text and images. Such a model could take over the responsibility of assessing whether an image looks like a text description, or vice versa, at any point in the pipeline where a human artist would do the same task. All of the above approaches can be applied in an iterative fashion over subsets of the search space, gradually limiting the range of possible values.

\subsection{Automation vs. Autonomy}

% While we can imagine a system being creative in its own right and productive out of its own motivation, this framework also aims to elevate the human artist to the role of director and to provide the possibility to guide a generative system through instructions and interventions.
While we have primarily focused on increasing a system’s creative autonomy through automation, our framework does not grant a system as much autonomy as to enable it to act entirely independently in response to its own motivations \citep[cf.][]{Guckelsberger2017a}. A system within our framework would remain inactive until engaged with. 
% The source of an intention, motivation or goal for creation thus remains with the human
Such engagement can range from a stimulus through available sensors, e.g.~cameras, microphones or heat sensors, to a text or image prompt or an entire inspiring set \citep{ritchie2007empirical}, to more precise and detailed instructions. In any case, this choice of input channel and sensibility has to be taken by a human and is not a target in our framework. 
% A human director might want to query a given system with a text prompt, in the style of a generative search engine.

We further assume the choice of generated media (image, audio, text, video, etc.) to be made by a person prior to building a system. Naturally, it is not difficult to imagine a setup in which this choice, too, becomes part of the pipeline.
Going one step further in autonomous automation, our framework and its targets make it possible to devise a generative system which produces automated generative pipelines. In fact, it might be possible for a generative system to generate itself, much like a general-purpose compiler that compiles its own source code. This self-referential generation has similarly been proposed in work on automated process invention \citep{charnley2014flowr}.

\subsection{Targets for Automation}

Below we define and discuss the many tasks and decisions that are part of a generative \ac{DL} pipeline in an artistic setting and which can be automated within our framework.
Wherever applicable, we explain how a target relates to concepts of \ac{AutoML} and \ac{CC}.

The following subsections identify individual targets for automation. The complete process is illustrated as a sequence of steps in figure~\ref{fig:pipeline}. As per this diagram, we organise the steps into three stages: (i) a \emph{preparation} stage to gather relevant materials (ii) a \emph{configuration} stage, where the models, training regimes and parameters are tuned to produce valuable output, and (iii) a \emph{presentation} stage where the user deploys a final model and curates the output.
The first target (selecting a pre-trained model) is optional and can be skipped in order to start from scratch instead. In this case, we begin with data preparation and curation.

\subsubsection{Pre-trained model (optional)}

It might not be necessary to train a network from scratch if an appropriate pre-trained model is available, especially when a quick system setup is desired. A list of pre-trained models, tagged with keywords associated to their generative domain, could provide a knowledge base for a system to select, download and deploy a model.
This can either be directly put to use, in which case the system could immediately skip to evaluating the model, or it can be fine-tuned on a smaller set of data. Such additional fine-tuning could be dependent on the outcome of the pre-trained model’s evaluation. Only if the pre-trained model’s output is not satisfactory would it have to be further optimised or de-optimised. 
Working with a pre-trained model has implications for the subsequent choices of the network architecture, training scheme and loss function.

\subsubsection{Data preparation and curation}

This preparation step includes the acquisition, cleaning, augmentation and transformation of data samples, akin to data preparation in \ac{AutoML}. 
% As the preparation of data prior to training usually requires a lot of manual adjustments, it provides many opportunities for automation. 
Starting with the data collection task, we consider different data sources from which a system could select. Drawing on existing data sets, such as an artist’s private data collection, can introduce important desirable biases and ensure high quality output. In contrast, scraping samples from the internet could contribute to the generation of surprising results. Additional pre-trained generative models can provide a source for synthesised data in large quantities.

An important addition to the pre-processing is data curation, in contrast to simple cleaning. Rather than filtering out noisy samples, for artistic purposes it can be desirable to add ‘noise’. To this end, it is not uncommon in an artistic context to mix multiple data sets. In this additional step, the system thus further refines the data set, similar to an artist adding or removing individual samples, which can influence the qualities of the system’s final output. This is an opportunity for iterative improvements and for \emph{alternating co-creativity} \citep{kantosalo2016modes}, given that the system both generates and evaluates.
Automation in the cleaning and curation tasks can be achieved, e.g.~in the image domain, by employing other computer vision or contrastive language-image models.

\subsubsection{Network architecture and training scheme}

This target for automation defines the choice of possible architectures (e.g.~\ac{GAN}, \ac{VAE}, Transformer), which could include non-neural methods. \Acf{NAS} in \ac{AutoML} is concerned with finding optimal combinations of basic building blocks of artificial neural networks in terms of performance on a classification or regression task, an immensely difficult optimisation problem. We recommend in our framework to instead select from tried-and-tested architectures, only altering parts of the architecture with a direct influence on the output, e.g.~the number of upsampling convolutions which determine the final output image size.

The training scheme is largely influenced by the choice of architecture. In the case of \acp{GAN}, the training scheme includes the choice of whether to train the discriminator and generator networks in parallel or consecutively, and how many individual optimisation steps to perform for either.

\subsubsection{Loss function}

The formulation of the basic loss term is highly dependent on a model’s training scheme and constitutes the the minimum requirement for successful training. However, additional loss terms can change or supplement the basic term for further refinement of the training objective. As a central part in guiding the model parameter optimisation process, any modification to the loss terms will strongly impact the modelled distribution and consequently the system’s output. In other contexts, methods have been proposed for the automatic invention of objective functions \citep{colton2008automatic}. These could provide a starting point for adapting the approach to the constraints of loss functions in generative \ac{DL}.

\subsubsection{Optimisation algorithm}

The selected algorithm will be responsible for adjusting a model’s parameters through error correction informed by the gradient of the loss function. This choice can potentially have an influence on the system’s output, as it is responsible for finding one of the potentially many local minima in the loss landscape. As it determines whether convergence can be reached at all, this decision can ultimately make or break the success of the training process. It can further largely influence convergence speed and be critical in time-sensitive setups. The choice of optimisation algorithms might be limited by the previous selection of network architecture and corresponding training scheme.

\subsubsection{Hyper-parameter tuning}

Optimisation of batch size, learning rate, momentum, etc. can be achieved via \ac{AutoML} methods, and there is much active research in this area.

\subsubsection{Model selection and evaluation}

% Stopping criteria
From all the possible models, the best one has to be selected in accordance with given criteria relevant to the task at hand. As the training process is essentially a succession of gradual changes of model parameters over time, this task is equivalent to identifying the right moment to stop training. 
% Selection criteria
Additionally, and in order not to lose previous training states, model checkpoints can be saved along the way as training progresses and whenever model evaluation satisfies given criteria. After training is finished, the best model has to be selected from all candidate checkpoints. In standard ML projects, this would normally be done with respect to the primary concern of predictive accuracy. But in generative projects, other considerations may include how surprising the outputs are, synthesis speed (for tool or real-time uses) and coherence of the results. 
%\notes{Explain how evaluation metrics can be selected or criteria be invented}
Such criteria could be employed in a weighted sum of metrics, where the system can give more or less emphasis to individual terms.
% and add its own invented metrics
This would allow the combination of standard metrics like \ac{FID} in the image domain for general output fidelity with a measure for sample similarity compared to a reference sample(s), inspiring set or text prompt via a contrastive language-image model.

% \begin{itemize}
%     \item Define stopping criteria
%     \item Or when to save model checkpoints
%     \item Validation metric corresponding to intention/goal
% \end{itemize}

% % Evaluation
% \begin{itemize}
%     \item Probe the model to understand whether it has successfully captured the desired domain and style of data. 
%     \item Check to see that there is sufficient diversity in the model output, to ensure that mode collapse and other common failure cases have not occurred. 
%     \item Might have to go back, adjust the data curation and re-train
% \end{itemize}

\subsubsection{Output curation}

Having obtained a successfully trained model, we want a system to reliably produce high-quality output. While efforts in previous steps were aimed at refining the model which is at the core of the generative process, this final automation target aims to raise the system’s overall output quality.
Two approaches come to mind: filtering and search.
In the former, a system could select those samples from a large batch of model outputs that rank highest against a given metric.
In the latter, the system could search for vectors directly in a model’s latent space via one of the various methods we have outlined in the section above on approaches to search problems.
The evaluation measure, as before, could be the similarity of samples compared to a set of reference samples, an inspiring set or a text prompt via a contrastive language-image model.

\section{An Illustrative Example}

\begin{figure}[t]
  \begin{minipage}[c]{0.215\textwidth}
    \includegraphics[width=\textwidth]{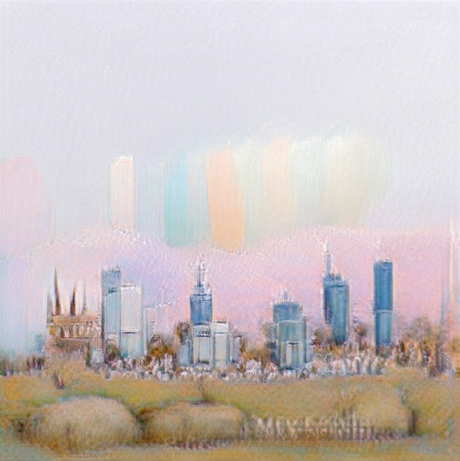}
  \end{minipage}\hfill
  \begin{minipage}[c]{0.23\textwidth}
    \caption{Image generated by the \emph{Big Sleep} Colab notebook for the prompt ``The Melbourne skyline in pastel colours''. Note the appropriate presentation of content and style, and additional pastel strokes in the sky as an unprompted innovation.} 
    \label{fig:melbourne}
  \end{minipage}
\end{figure}

In early 2021, a generative deep learning Colab notebook \citep{bisong2019google} called the \emph{Big Sleep} was shared online \citep{murdock2021big}. It allows for text-to-image generation \citep{agnese2020survey}, effectively visualising a user-given text prompt, often with innovative content and design choices, as per the example in figure \ref{fig:melbourne}. This is an instance of an artistic deviation from the standard pipeline, where CLIP \citep{radford2021learning} is used to evaluate a generated image w.r.t. a given text prompt, driving a gradient-based search for latent vector inputs to a generative model called BigGAN \citep{brock2019large}. We use this setup as an example to identify the following places where automation could be introduced, based on our framework. We highlight concrete techniques and references for automation from the literature.

\vskip 0.05in
\point{In terms of \textbf{pre-trained model selection}, numerous people have substituted BigGAN with other GAN generators. This creative responsibility could be automated, with the system choosing from a database of GANs and installing new ones into the notebook. }

\vskip 0.05in
\point{In terms of \textbf{data preparation and curation}, users often choose imaginative text prompts, as the notebook often produces high quality, surprising results for these. This could be substituted, for example, with automated fictional ideation techniques \citep{llano2016automated}.}

\vskip 0.05in
\point{\citet{murdock2021big}, the notebook programmer, innovated in \textbf{loss function definition}, employing patches from generated images rather than the entire image to evaluate its fit to the prompt. Various image manipulation routines could be automatically tested within loss function calculations from a library, with the system automatically altering the notebook at code level.}

\vskip 0.05in
\point{As described in \citet{colton2021generative}, in some circumstances where multiple images are being generated simultaneously, increasing the learning rate can help searches fail quickly. Such \textbf{hyper-parameter tuning} could be automated using standard AutoML techniques, guided by requirements on acceptable search successes and output image quality.}

\vskip 0.05in
\point{In terms of \textbf{model selection and deployment}, we can imagine models being used as creative web-services \citep{veale2013creativity}, with higher-level \ac{CC} systems accessing text-to-image generators in a variety of projects.} 

\vskip 0.05in
\point{There has been an explosion of human usage of notebooks like the \emph{Big Sleep}, with attendant \textbf{output curation} via cherry picking results for posting on social networks and in blogs. This would be an ideal target for automation with systems using CLIP and other techniques to evaluate images, also possibly inventing new aesthetic measures \citep{colton2008automatic}.}

%\notes{Describe how parts of the big sleep notebook pipeline could be extended with the help of our framework, in automation, organisational structure and higher level CC goals. 1) selection of pre-trained generator model, 2) definition of stop criteria, 3) Image input}

\vskip 1.1cm

\section{Discussion}

We have presented a framework for the specific purpose of automating manual tasks in a generative \ac{DL} pipeline for artistic projects. We adopt the core concepts of \ac{AutoML} and adjust them in two ways. First, we focus on generative \ac{DL} which differs in the type of learning task, in that it is concerned with modelling the distribution of a training set, rather than classification or regression. And second, we address the artistic usage of generative \ac{DL}, where more emphasis is given to the qualities of the generated output over the qualities of the model. The specialisation of our framework inversely limits its generalisability in the same ways. On the one hand, there might be artefact-driven applications of generative \ac{DL} within or outside \ac{CC} that we have not considered. On the other hand, our framework is not generally applicable to generative approaches in \ac{DL} due to its special emphasis on artistic uses. Its focus on generative \ac{DL} further limits its validity for other generative modelling methods.

% \notes{Further limitations in automation}

\subsection{Automation and Creative Autonomy}
We have previously analysed the close relationship between the \emph{automation} of generative \ac{DL} systems and the central \ac{CC} goal to increase a system's \emph{creative autonomy}~\citep{jennings2010developing,mccormack2019autonomy, Guckelsberger2017a} by granting it more \emph{creative responsibilities}~\citep{colton2008creativity}. Here, we complement this \emph{a priori} analysis based on knowledge of our concrete automation pipeline. The aim is to understand to which extent our proposed pipeline already enables facets of creative autonomy, and how \ac{CC} insights on creative autonomy could be used to advance it in future work.

%We analyse the intricacies of this relationship to understand to which extent our proposed automation pipeline already enables facets of creative autonomy, and how it could be developed further in future work.

Automation is necessary for creative autonomy, but the opposite does not hold: while a fully automated generative \ac{DL} system might still exactly follow user-prescribed goals, an autonomously creative system has the \enquote{freedom to pursue a course independent of its programmer’s or operator’s intentions} \citep{jennings2010developing}. This firstly requires the system to autonomously \emph{evaluate} its creations, which is satisfied by any system that can be considered \emph{creative} \citep{ventura2016mere}. In addition, an \emph{autonomously creative} system must be capable of autonomous \emph{change}, i.e.~initiating and guiding \enquote{changes to its standards without being explicitly directed when and how to do so} \citep{jennings2010developing}. To prevent trivial implementations of these capabilities, \citeauthor{jennings2010developing} requires them to not exclusively rely on random decisions.

% \notes{Others: note that avoiding random decisions can also be a means to avoid inefficiency in ML automation, e.g. in hyper-parameter tuning - you might be proposing some of this already in your pipeline.}

% \notes{Others: check if use of 'goals' here is consistent with terminology in your pipeline - probably needs changes!} 

To assess how much our pipeline realises creative autonomy, we can draw on various \ac{CC} approaches to enhancing autonomy in computational systems. For instance, \citet{colton2009seven} proposes \enquote{repeatedly asking ourselves: what am I using the software for now? Once we identify why we are using the software, we can \textelp{}~write code that allows the software to use itself for the same purpose. If we can repeatedly ask, answer and code for these questions, the software will eventually \textelp{}~create autonomously for a purpose, with no human involvement}. 
Our framework provides various candidate targets to perform such a gradual elevation of a generative \ac{DL} system. 

For the evaluation of a concrete system built under our framework, we consider the FACE model \citep{colton2011computational,pease2011computational} an adequate evaluation tool. In this evaluation model, systems are described in terms of the creative acts they perform. Such an analysis allows for the identification of newly added creative responsibilities through automation.

\citet{linkola2017aspects} follow a more constrained approach and, as part of a larger agenda to realise meta-creativity in \ac{CC}, propose that creative autonomy requires \emph{artefact-}, \emph{goal-} and potentially \emph{generator-awareness}, realised through operators of \emph{(self-) reflection} and \emph{(self-) control} which closely match Jennings’ \citeyearpar{jennings2010developing} requirements for evaluation and change. 
Whether a system built within our framework satisfies these definitions depends on the extent to which it is granted responsibilities in the form of automating decision making for targets identified in the framework. We demonstrate this based on extensions to a non-automated generative \ac{DL} system.
% self-reflection/evaluation: specific loss function
% self-control/change: back-propagation and error correction
Such a system can be considered to have some generator-awareness due to the role of its loss function (self-reflection), and its adjustments of own parameters through error correction methods like back-propagation (weak self-control). 
A system’s control over changes to its generator can be increased from weak to strong within our framework, through the automated manipulation of network architecture or selection of a pre-trained model.
Further putting a system in charge of its loss function within our framework (strong control) affords it goal-awareness and consideration as autonomously creative, if it is capable of modifying the loss function in response to its evaluation of generated output.

Crucially, more radical forms of creative autonomy do not eliminate co-creation, i.e.~cut ties with the system user entirely, but facilitate different forms of interaction. To really become independent of its designer, a system must not be isolated but interact with critics and creators that shape its evaluation and changes~\citep{jennings2010developing}. A fully creatively autonomous system might refuse the will of its interaction partner \citep{jennings2010developing,Guckelsberger2017a}, but we believe that this holds a promise for innovative artistic collaborations between people and computational systems, connecting artistic practices in generative \ac{DL} with the philosophy and goals of CC.
%Increasing creative autonomy requires shifting creative responsibility to the system, yet allows people to interact with the system towards shared goals. 
%This aligns with our framework’s goal to connect artistic practices with the philosophy and goals of CC, opening up a world of creative autonomy for generative \ac{DL} projects.

\subsection{Future Work}

In this formulation of our framework we have only briefly mentioned automation of creative responsibilities via the usage of \ac{ML} models. The possibility of training or deploying multiple models in the same system enables the addition of organisational structures to our framework, in which we think of individual models as agents in a multi-agent system.

To use our framework in co-creative applications, augmenting a system with the ability to communicate its adjustments and intentions would be especially beneficial. 
% This ties in with the higher goal of explainable \ac{CC}
Moreover, to address our framework’s limitations, further work is needed to consider applications which use generative \ac{DL} but are not artistically focused. This could potentially inform a more general automated \ac{ML} framework, which would further benefit from more formal definitions.

We plan further study of the ways in which deep learning researchers, practitioners and artists work with generative systems, in particular where they have, and could, add levels of automation, via analyses such as the illustrative example above. Some of the techniques that artists apply, such as data set curation and iteration, as well as the selection of generated outputs, are promising avenues for automation and require further investigation. 
We further plan to put our framework to use in applied projects. Through this, we aim to provide demonstrative examples of how some of the challenges in automation can be tackled and to show the surprising results that automation can afford. For the evaluation of such demonstrative examples we plan to draw from the FACE descriptive model of creative acts \citep{colton2011computational,pease2011computational}.
% As with \ac{AutoML} for standard machine learning projects, we want to relieve people of some of the burden of data management, training/deploying models and curating outputs. However, we also want to expose ML researchers to the many benefits in terms of surprising, enlightening and inspiring outputs and processes to be gained by sharing creative responsibilities with generative software. 

% — Organisational structures, 1) with models as agents (endogenous connections in Linkola et al.), and 2) in between generative systems
% — Further autonomy: 1) choice of generative medium, 2) input stimulus (channel, type, sensitivity), essentially what would a system react to?, 3) intrinsic motivation, why would a system act autonomously at all?
% – Explanation of the system of its process and decisions; helpful for co-creative interaction with human artist

% Opportunities highlighted by the framework:

% — Automatic invention of loss terms specific to the constraints of NNs

\section{Acknowledgements}
We thank our reviewers for their helpful comments. Sebastian Berns and Terence Broad are funded by the EPSRC Centre for Doctoral Training in Intelligent Games \& Games Intelligence (IGGI) [EP/L015846/1, EP/S022325/1]. Christian Guckelsberger is supported by the Academy of Finland Flagship programme \enquote{Finnish Center for Artificial Intelligence} (FCAI).

% Define acronyms
\begin{acronym}
    \acro{AutoML}[{AutoML}]{automated machine learning}
    \acro{CC}{computational creativity}
    \acro{CLIP}{Contrastive Language-Image Pretraining}
    \acro{DL}{deep learning}
    \acro{FID}{Fréchet Inception Distance}
    \acro{GAN}{generative adversarial network}
    \acro{ML}{machine learning}
    \acro{NAS}{neural architecture search}
    \acro{VAE}{variational auto-encoder}
\end{acronym}

% \newpage

\bibliographystyle{iccc}
\bibliography{references}

\end{document}